\documentclass[lettersize,journal]{IEEEtran}
\usepackage{amsmath,amsfonts}
\usepackage{algorithmic}
\usepackage{algorithm}
\usepackage{array}
\usepackage[caption=false,font=normalsize,labelfont=sf,textfont=sf]{subfig}
\usepackage{textcomp}
\usepackage{stfloats}
\usepackage{url}
\usepackage{verbatim}
\usepackage{graphicx}
\usepackage{cite}
\usepackage{xcolor}
\usepackage{xspace}
\usepackage{multirow}
\usepackage{threeparttable}

\newcommand*{\revise}[1]{\textcolor{black}{#1}}

\newcommand*{\jdw}[1]{{\textcolor{black}{#1}}}
\newcommand{\sysname}{$\text{Farm-LightSeek}$\xspace}
\begin{document}

\title{\sysname: An Edge-centric Multimodal  Agricultural IoT Data Analytics Framework with Lightweight LLMs}

\author{\IEEEauthorblockN{Dawen Jiang, Zhishu Shen,
 Qiushi Zheng, Tiehua Zhang, Wei Xiang, and Jiong~Jin}

\thanks{Dawen Jiang and Zhishu Shen are with the School of Computer Science and Artificial Intelligence, Wuhan University of Technology, Wuhan, China.}
\thanks{Qiushi Zheng and Jiong Jin are with the School of Engineering, Swinburne University of Technology, Melbourne, Australia.}
\thanks{Tiehua Zhang is with the School of Computer Science and Technology, Tongji University, Shanghai, China.}
\thanks{Wei Xiang is with the School of Computing, Engineering and Mathematical Sciences, La Trobe University, Melbourne, Australia.}
\thanks{Corresponding author: Zhishu Shen (z\_shen@ieee.org) and Jiong Jin (jiongjin@swin.edu.au)}

}

\markboth{Journal of \LaTeX\ Class Files,~Vol.~14, No.~8, August~2021}%
{Shell \MakeLowercase{\textit{et al.}}: A Sample Article Using IEEEtran.cls for IEEE Journals}


\maketitle

\begin{abstract}
Amid the challenges posed by global population growth and climate change, traditional agricultural Internet of Things (IoT) systems is currently undergoing a significant digital transformation to facilitate efficient big data processing. 
\jdw{While smart agriculture utilizes artificial intelligence (AI) technologies to enable precise control, it still encounters significant challenges, including excessive reliance on agricultural expert knowledge, difficulties in fusing multimodal data, poor adaptability to dynamic environments, and bottlenecks in real-time  decision-making at the edge. Large language models (LLMs), with their exceptional capabilities in knowledge acquisition and semantic understanding, provide a promising solution to address these challenges.} To this end, we propose \sysname, an edge-centric multimodal agricultural IoT data analytics framework that integrates LLMs with edge computing. This framework collects real-time farmland multi-source data (images, weather, geographic information) via sensors, performs cross-modal reasoning and disease detection at edge nodes, conducts low-latency management decisions, and enables cloud collaboration for model updates. The main innovations of \sysname include: (1) an agricultural ``perception-decision-action" closed-loop architecture; (2) \revise{cross-modal adaptive monitoring}; and (3) a lightweight LLM deployment strategy balancing performance and efficiency. 

\jdw{Experiments conducted on two real-world datasets demonstrate that \sysname consistently achieves reliable performance in mission-critical tasks, even under the limitations of edge computing resources. }This work advances intelligent real-time agricultural solutions and highlights the potential for deeper integration of agricultural IoT with LLMs.

\end{abstract}

\begin{IEEEkeywords}
Smart agriculture; Internet of Things; large language model; edge computing; multimodal data analytics; knowledge distillation
\end{IEEEkeywords}

\section{Introduction}
\IEEEPARstart{I}{ntelligent} agriculture is a revolutionary paradigm leveraging advanced Internet of Things (IoT) and artificial intelligence (AI) technologies to modern farming practices such as smart irrigation and pest management~\cite{PaganoIoT23}. By deploying IoT sensors, field robots, and automated irrigation systems, the intelligent agricultural systems enable real-time environmental monitoring and precise control of farmland conditions, significantly improving resource efficiency. This shift addresses the limitations of traditional agricultural systems that rely on manual labor and static decision-making, while enhancing productivity, sustainability, and operational precision.

The effective implementation of smart agriculture is inseparable from precise environmental monitoring and timely decision-making in agricultural management, demanding substantial expertise from practitioners. Current agricultural IoT systems still face challenges in integrating and processing multimodal heterogeneous data, such as images, time-series sensor data, and meteorological records, for real-time decision-making in complex and dynamic environments. Existing approaches mainly employ traditional machine learning methods like support vector machine (SVM) and random forest, or single-modality deep learning models like convolutional neural network (CNN), long short-term memory (LSTM)\cite{MUHAMMED2024103905},  however, these methods face critical limitations: (1) a lack of model generalizability and adaptability under dynamic environmental conditions, and (2) insufficient cross-modal semantic reasoning capabilities, hindering closed-loop automation from environmental perception to decision-making. Large language models (LLMs) provide an exceptional paradigm for addressing these issues. Trained on massive datasets, LLMs excel at knowledge acquisition and semantic understanding, functioning as ``domain-agnostic experts" capable of interpreting cross-disciplinary problems. By integrating LLMs with agricultural IoT domain knowledge, it is expected to develop dedicated ``smart brains" for individual farmhouses, enabling intelligent, granular production management and pioneering new paradigms for agricultural multimodal data fusion.

Traditional cloud-centric AI architectures, which centralize data processing in the cloud before relaying results to edge devices, face the issues including latency bottlenecks, privacy vulnerabilities, and prohibitive operational costs. Edge computing addresses these challenges by decentralizing computation, bringing processing and storage directly to the source of data generation. This transforms each independent farmhouse into self-sufficient ``computing hub", completing the comprehensive data analytics from information acquisition to decision-making locally. While edge computing reduces reliance on the cloud and minimizes latency by moving tasks closer to data sources, its constrained hardware capability struggles to support traditional LLMs. Existing small-scale LLMs reduce computational demands, but often degrade multimodal reasoning capabilities, which are vital for interpreting diverse agricultural data in smart agriculture.

\begin{figure}
    \centering
    \includegraphics[width=1\linewidth]{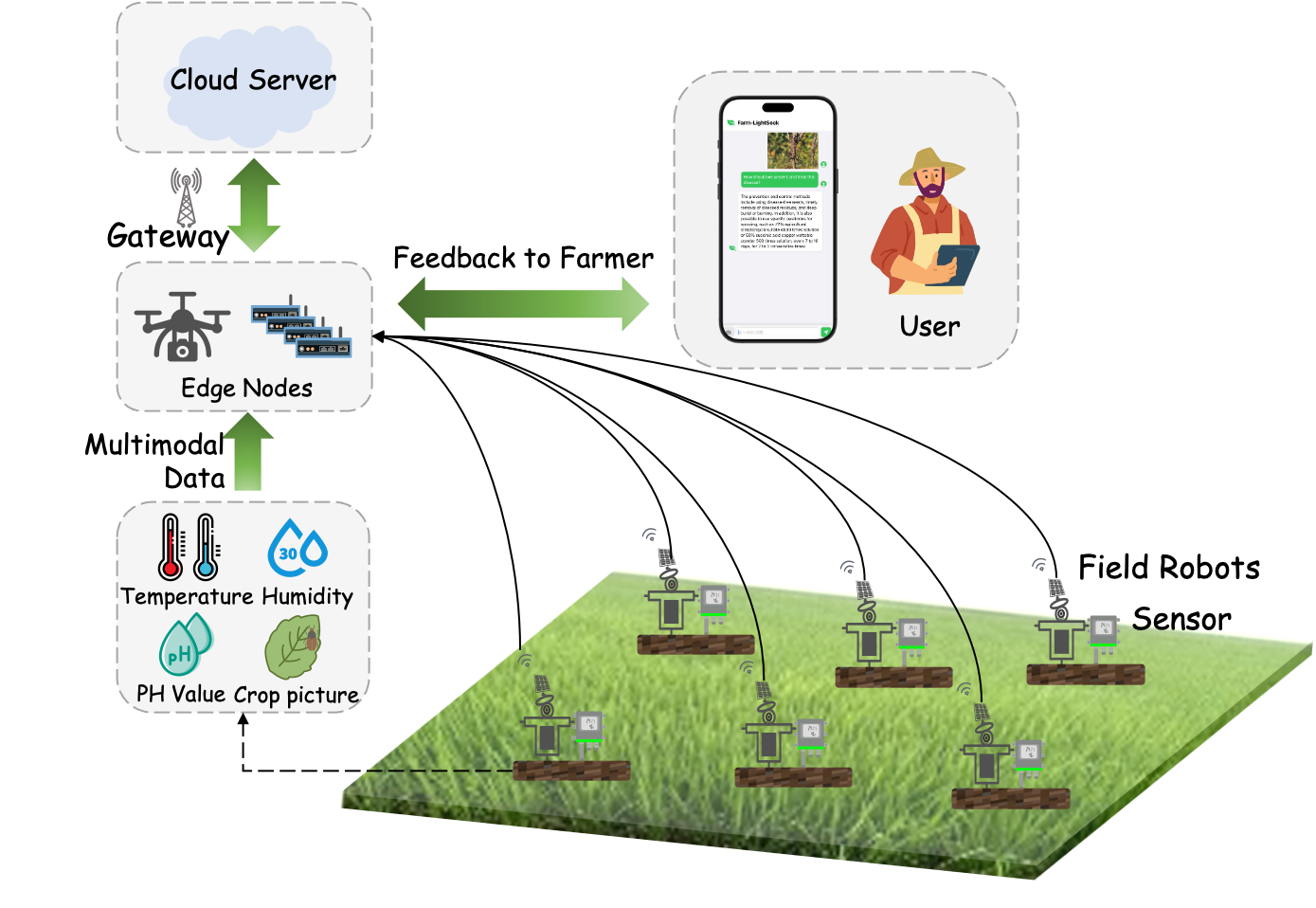}
    \caption{Multimodal agricultural IoT data management of \sysname.}
    \label{fig:framework_network}
\end{figure}

To this end, we propose \sysname, an edge-centric multimodal agricultural IoT data analytics framework that combines networking and lightweight LLMs. This framework aims to generate sustainable and accurate management recommendations through edge computing and multimodal data fusion in \revise{resource-constrained} agriculture environments. As shown in \figurename~\ref{fig:framework_network}, \sysname collects environmental multimodal information (including images, meteorological data, geographical location) in real time through mobile or fixed sensor devices such as drones and soil probes. We compress the multimodal LLM (MLLM) with a large number of parameters through a three-stage knowledge distillation so that it can be deployed on edge nodes. Edge nodes obtain management decisions through pest detection and recommendation generation based on MLLM, and feedback the results to users through the client or IoT devices in real time. During idle communication periods, edge nodes exchange the stored compressed data with cloud servers, facilitating timely updates to the model. We also verify the effectiveness of \sysname through the latest agricultural visual question answering (VQA) benchmarks and multi-round dialogue data. The main contributions of this work are in the following aspects:
\begin{enumerate}
    \item To the best of authors' knowledge, this is the first article that presents an edge-centric LLM-based agricultural IoT data analytics framework that achieves closed-loop management from data awareness to decision execution. 
    \item To support data analytics at \revise{resource-constrained} edge nodes, we develop a lightweight version of MLLM using advanced knowledge distillation techniques, reducing computational overhead while maintaining satisfactory inference accuracy.

    \item We conduct comprehensive performance evaluations to validate the reliability and efficiency of \sysname in real-world agricultural datasets using multidimensional indicators. 
   
\end{enumerate}

\section{Background}
\subsection{Edge-centric Agricultural IoT Systems}

Existing agricultural services can be categorized into environmental monitoring, smart irrigation, precision fertilization, and pest management. Traditional IoT systems involve collecting environmental data such as soil temperature, humidity, and light intensity through sensor networks, and leveraging cloud computing platforms for data analysis and decision support~\cite{PaganoIoT23}. 
However, the centralized approaches encounter challenges including excessive network latency and bandwidth demands, rendering them impractical for real-time applications such as pest alerts and precision irrigation. Shifting data processing function to edge nodes located in the vicinity of the user site enables real-time data analytics in resource-constrained agricultural environments. Recent progress in edge-centric frameworks facilitates the lightweight data analytics, allowing local IoT systems to improve resource efficiency to boost crop productivity~\cite{XuIEEE21}.  

\textbf{Environmental monitoring:} Edge computing supports on-device processing of data from distributed sensors and imaging devices. For instance, edge-deployed lightweight CNNs have been utilized to analyze multispectral images captured by drones for crop health assessment, enabling early detection of nutrient deficiencies or water stress without relying on cloud infrastructure. Similarly, edge nodes equipped with low-power vision processors have been integrated with soil sensors to monitor microclimate conditions and trigger localized alerts for frost or drought risks.  

\textbf{Smart irrigation:} Edge computing frameworks enhance water management by dynamically analyzing live soil moisture levels and meteorological conditions. Intelligent irrigation systems leverage federated learning algorithms to adaptively optimize water distribution across agriculturally diverse zones with varying needs. This approach minimizes resource waste, achieving efficiency improvements over traditional irrigation methods.

\textbf{Pest management:} Pest and disease management remains a critical challenge for edge computing-based agricultural IoT due to the need for rapid, high-resolution image analysis. Edge computing addresses this by enabling on-device execution of lightweight object detection models. For example, edge-deployed YOLOv5-Tiny models have been integrated into smart traps to identify and count pests in real time, reducing the reliance on cloud-based solutions. In disease monitoring, edge nodes process RGB and thermal images from handheld devices to segment diseased leaf areas using U-Net variants optimized for edge hardware. However, intelligent systems capable of automating pest control actions and providing comprehensive management recommendations remain underdeveloped, as most systems focus solely on detection rather than integrated management and lack authoritative datasets for response strategies.

Despite significant advancements, existing edge-centric IoT systems still encounter critical limitations. First, implementations depend on costly communication and computing infrastructure, making them economically critical for smallholder farmers. Second, interoperability challenges between edge platforms and conventional farming equipment continue to hinder widespread adoption. Third, the incorporation of agricultural domain expertise into edge-based AI models remains underdeveloped. To address these challenges, it is crucial to design a lightweight edge intelligence framework, establish cross-platform compatibility standards, and develop hybrid systems that integrate AI with expert knowledge to improve decision-making capabilities in smart agriculture.

\subsection{Agricultural Multi-modality Large Language Model}

LLMs like ChatGPT and DeepSeek exhibit strong zero-shot learning abilities, solving diverse tasks through text-based prompts without task-specific training. However, their dependence on single-modality input (text) restricts their applicability in contexts demanding multimodal analysis, such as agricultural scenarios involving field imagery, spectral data, or sensor streams. While LLMs can generate detailed textual descriptions of pests or diseases, they lack the capacity to directly process visual or sensor-derived data for tasks such as crop disease diagnosis or irrigation optimization, a key limitation in agricultural IoT systems.

MLLMs address this limitation by integrating vision, language, and other modalities through architectures such as cross-modal attention and contrastive pre-training. Recent advances emphasize efficient multimodal fusion and instruction-driven adaptability. For example, LLaVA~\cite{LLaVa} uses the synthetic data generated by GPT-4 to align visual and textual features through instruction tuning. Its two-stage training—first fine-tuning on filtered image-text pairs and then refining with GPT-4-generated instructions—exemplifies the shift toward data-efficient multimodal reasoning. Uni-MoE~\cite{LiTPAMI25} introduces a mixture-of-experts (MoE) architecture to dynamically activate modality-specific pathways, enabling efficient fusion of text, images, and time series sensor data with minimal computational overhead. OmniVL unifies video, audio, and text through temporal-aligned cross-modal attention. 

The versatility of MLLMs has been validated in cross-domain applications. In legal analysis, Lawyer LLaMA enhances legal reasoning and accuracy through specialized training while maintaining generalizable capabilities applicable to other fields. In natural science, GeoCLIP\cite{GEOCLIP} aligns satellite imagery with geospatial texts to predict climate patterns, demonstrating the potential for environmental monitoring. However, in agricultural IoT, multi-mode data fusion is very important. Due to the lack of computing resources in agricultural environments, it is difficult to deploy large-scale MLLM, and lightweight MLLM has not been fully developed. Current systems mainly rely on single-mode LLM to obtain text data, so it is difficult to synthesize heterogeneous inputs, such as multispectral images and sensor readings. Current systems primarily rely on single-modality LLMs for textual data, struggling to synthesize heterogeneous inputs such as multispectral images and sensor readings. Our work focuses on leveraging the multimodal fusion capabilities of MLLMs to transform raw agricultural data into precise field-deployable recommendations, addressing the unique challenges of smart agriculture.  

\subsection{Knowledge Distillation (KD)}
KD aims to transfer knowledge from complex large teacher models to lightweight student models, enhancing the student models' performance with fewer parameters and computational costs. Traditional KD methods utilize soft logits from the teacher as supervision signals. Subsequent advancements, such as FNKD~\cite{FNKD}, demonstrated that mimicking the teacher’s intermediate features further improves classification accuracy. DGKD~\cite{DGKD} improved the student model’s predictions, by integrating guidance from multiple teachers.  

\textbf{KD for LLM:} With the rise of LLMs such as ChatGPT, more and more research is focused on improving the accuracy of models. As a result, the size of the model continues to increase, limiting the efficient deployment of LLMs in resource-constrained scenarios. As a result, some researchers have begun to apply knowledge distillation techniques to LLMs. For instance, MiniLLM\cite{minillm} and DistiLLM\cite{distillm} optimize distillation by proposing novel loss functions, reverse Kullback-Leibler Divergence (KLD) and skew KLD, to mitigate the student’s overemphasis on the teacher’s long-tail output distributions. Adaptive strategies, such as balancing KLD and reverse KLD losses, further refine this process. To enhance reasoning capabilities, methods like TinyLLM~\cite{tinyllm} leverage multiple teacher models, while others utilize the Chain-of-Thought (CoT) ability of large LLMs to model causal relationships and generate enriched training data.  

\textbf{KD for MLLMs:}  Extending KD to MLLMs, LLaVA-MoD~\cite{llavamod} introduces structural and training innovations. By integrating an MoE architecture into the student MLLM, the model enhances its capacity to handle heterogeneous data (e.g., images, text). During training, standard KLD aligns the student’s response logits with the teacher’s outputs, while a preference distillation process reduces hallucinations by refining the student’s judgment through human-like preference learning. These advancements underscore the potential of KD in democratizing MLLMs for scenarios requiring efficient, yet accurate, multimodal reasoning.  LLaVA-KD~\cite{LLaVA-KD} focuses on optimizing training schemes and developing multimodal distillation strategies to effectively and efficiently improve the performance of existing small-scale MLLMs in a single-teacher model.

\section{\sysname Framework}

\begin{figure*}[tb!]
    \centering
    \includegraphics[width=0.98\linewidth]{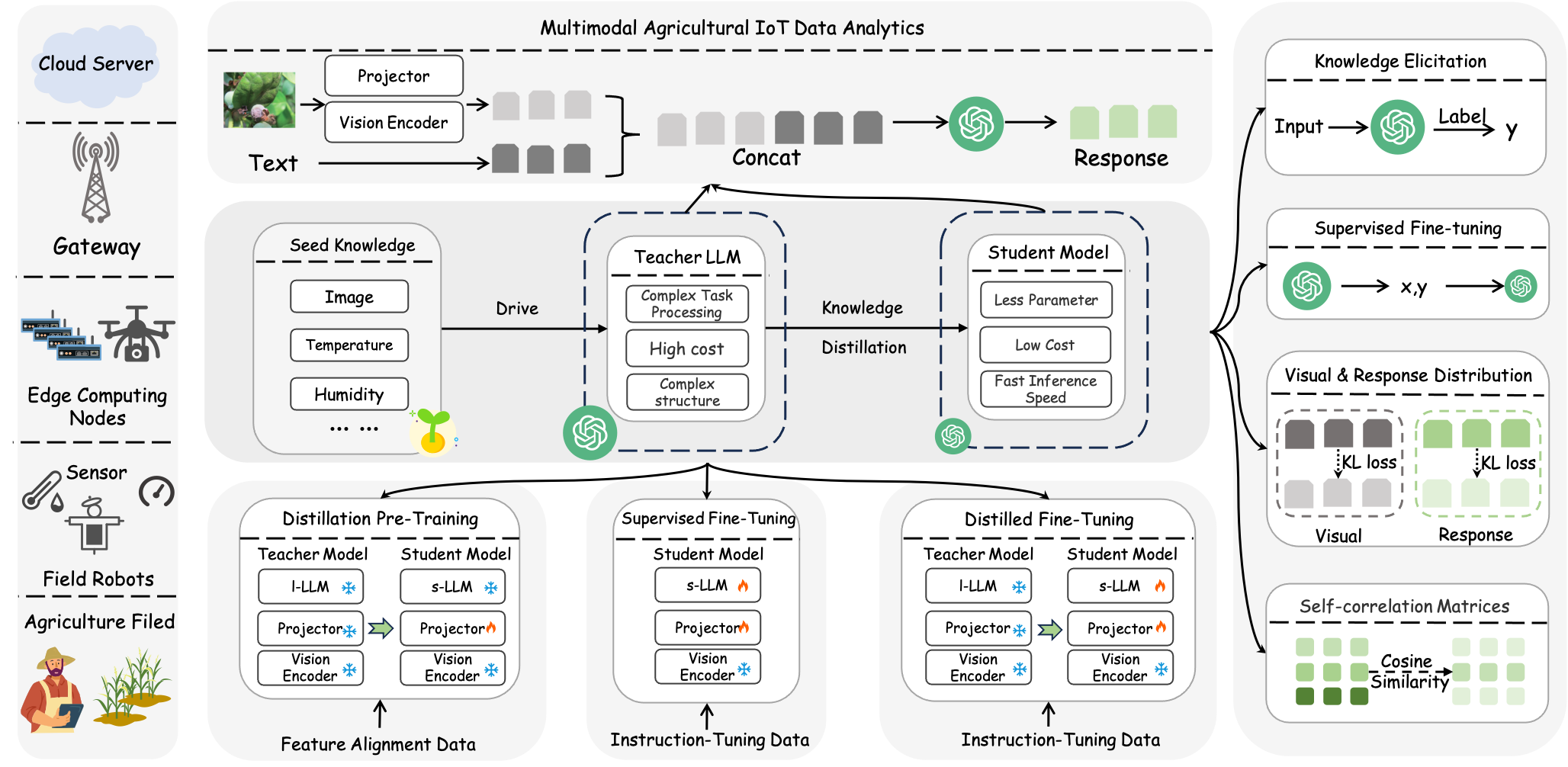}
    \caption{Overview of the proposed framework \sysname.}
    \label{fig:framework}
\end{figure*}

Despite the zero-shot learning capabilities of LLMs, deploying them in agricultural applications remains challenging due to constrained communication and computational resources across vast farmlands, as well as the complexity of processing multimodal agricultural data. Traditional agricultural IoT systems, which rely on sensor networks and manual monitoring, often fail to enable high-precision decision-making in real time. To address these limitations, we explore the application of MLLMs in agriculture and propose \sysname, an edge-centric multimodal IoT analytics framework designed for agricultural environments using lightweight LLMs.  

\figurename~\ref{fig:framework} illustrates the overall of \sysname in terms of multimodal agricultural IoT data analytics. This framework leverages mobile devices (e.g., field robots, fixed sensors) to capture multimodal data such as environmental parameters (humidity, temperature, light intensity) and crop imagery. A knowledge-distilled, lightweight MLLM deployed on edge nodes like agricultural drones, aligns these heterogeneous image-text inputs to generate precise management recommendations in real-time. \revise{Potential edge nodes for data processing include the NVIDIA Jetson NANO 4GB Developer Kit for drones and  Qualcomm QCS610 for on-device RGB/thermal image processing.} Edge nodes communicate with cloud servers through gateway intermediaries, where the cloud serves as a centralized hub for data storage and continuous model refinement. The updated models are then distributed back to the corresponding edge nodes for further deployment.

\subsection{Multimodal Agricultural Language Model}~\label{sec:3a}
To address multimodal challenges in agricultural IoT applications, we employ LLaVA, a vision-language integration model to diagnose crop pests/diseases and produce actionable agricultural insights. LLaVA integrates visual and linguistic understanding through a unified visual-language architecture. Its core structure comprises three key components: (1) a CLIP-ViT visual encoder that converts images to patch-based embeddings, (2) a language model for text processing and generation, and (3) a projection layer that aligns visual features with the token space of the language model. The projection layer, typically implemented as a lightweight linear or MLP-based module, bridges the modality gap by mapping visual embeddings into word-like tokens, enabling seamless fusion of image and text inputs. For example, a 336$\times$336 pixel image is split into 16$\times$16 patches, encoded by CLIP-ViT into 576 tokens, which are then projected into 256-dimensional vectors compatible with the hidden states of the language model.  

During training, LLaVA leverages multimodal instruction datasets synthesized by GPT-4, comprising image-text pairs annotated with task-oriented directives (e.g., ``Describe the pest damage in this crop image”). The workflow involves the following stages:

\subsubsection{Pretraining} Image-text pairs are used to align visual and textual features. The visual encoder and projection layer are trained to minimize the discrepancy between image embeddings and corresponding text descriptions. \revise{For numerical data such as pH values, we convert them into text-based representations and embed them into descriptive prompts, such as ``Current sensor data: pH=5.8, temperature=28°C. Analyze whether the crops in this image exhibit abnormalities."}  

\subsubsection{Instruction Tuning} GPT-4 synthesizes diverse instruction-response pairs (e.g., Q\&A, reasoning tasks) from raw image captions. For example, an image of wilted leaves could be paired with the instruction ``Diagnose the crop health problem" and a response detailing the signs of fungal infection. The model learns to generate context-aware answers by autoregressively predicting text tokens conditioned on both visual and textual inputs.  

\subsection{Edge-deployable Model} 
To address the challenges of deploying large models on resource-constrained devices, we conducted the following work:
(1) Inspired by TinyLLaVA\cite{TinyLLaVA}, we reconstructed the original architecture of LLaVA by replacing its language model with Qwen2.5-0.5B, thereby obtaining a small-scale MLLM. \revise{Compared to existing open-source models, Qwen2.5-0.5B exhibits enhanced competitiveness in tool invocation capabilities and long-context processing.} (2) Since training compact MLLMs with native strategies often yields suboptimal performance, we design a three-stage KD pipeline inspired by \cite{LLaVA-KD} to transfer capabilities from a larger MLLM to our lightweight version, aligning its performance with the large-scale model. The details of this distillation framework are outlined below:

\subsubsection{Distillation Pre-Training (DPT)} 
\revise{As illustrated in \figurename~\ref{fig:framework}, in this stage, we freeze the visual encoder and LLM of the small MLLM (s-MLLM), and only optimize the projector. The student model is trained to acquire the teacher model’s efficient encoding capabilities for agricultural images by minimizing the Kullback-Leibler (KL) divergence between their visual and textual outputs. For instance, in leaf disease spot detection, the student model demonstrates enhanced precision in mapping image features (e.g., brown lesion morphology and chromatic characteristics) to textual descriptions (e.g., ``potato late blight"). Concurrently, optimizing the cosine similarity between visual feature autocorrelation matrices improves the output quality of the student model's visual features, enabling it to assimilate the teacher model's proficiency in modeling local feature correlations.}

\subsubsection{Supervised Fine-Tuning (SFT)} 
\revise{As illustrated in \figurename~\ref{fig:framework}, based on the DPT, this stage adopts the conventional SFT methodology, in which the visual encoder is frozen, and the projector and the small language model (s-LLM) are jointly trained using high-quality dialogue data. This process endows the model with reasoning capabilities and instruction-following proficiency.}

\subsubsection{Distilled Fine-Tuning (DFT)} 

\revise{As shown in \figurename~\ref{fig:framework}, in the DFT stage, knowledge transfer is further deepened to enhance reasoning capabilities for complex agricultural tasks. During this stage, the visual encoder remains frozen, with optimization concentrated on the projector and s-LLM, while strategies from the DPT stage are reused. By aligning the response distributions, visual distributions, and visual feature auto-correlation matrices between the teacher model and the student model, we effectively transfer the complex reasoning abilities and visual representation capabilities of the l-MLLM to the s-MLLM.}

The training process for large-scale MLLMs outlined in Section~\ref{sec:3a} serves as the performance upper bound of the small model. Our three-stage distillation strategy enables these smaller models to closely approximate the capabilities of the large model.

\section{Experiments}
\subsection{Experimental Settings}

\jdw{As demonstrated in Table~\ref{tab:comparison}, 
the efficacy of the proposed \sysname framework is validated through two real-world datasets\revise{\cite{Agri_LLaVA}}: an agricultural feature alignment dataset and an agricultural instruction-tuning dataset. The former serves as a foundational multimodal corpus, facilitating the alignment of visual and textual representations of agricultural pests and diseases. The latter constitutes a high-quality conversational corpus specifically designed for domain-specific multimodal dialogue generation. Both datasets are designed to enhance the model's capability in interpreting and addressing agricultural challenges by integrating structured knowledge and expert-validated dialogues, thereby ensuring the model's rigorous and reliable performance.}

\begin{table*}[]
\caption{The details of two datasets}\label{tab:comparison}
\centering
\renewcommand{\arraystretch}{1.3} 
\begin{threeparttable}
\scalebox{0.88}{
\begin{tabular}{c|cc}
\hline
Name      & Feature Alignment Data                                                                                                                                                                                                   & Instruction-Tuning Data                                                                                                                                                               \\ \hline

Scale           & Approximately 400,000 samples                                                                                                                                                                                            & 6,000 high-quality samples                                                                                                                                                            \\ \hline
\revise{Categories}	& \revise{221 types (e.g., late blight, powdery mildew, flea beetles) }
& \revise{5,813 images of infected crops paired with domain knowledge}
\\ 
\hline
Source          & 16 publicly available datasets (e.g., IP102)                                                                                                                                                                             & Websites such as  CAAS Pests and Diseases Database{\tnote{1}}                                                                                                                                                           \\ \hline
Purposes    & \multicolumn{1}{l}{\begin{tabular}[c]{@{}l@{}}-Enable the model to recognize image features and classify pests and diseases.\\ -Enter symptom knowledge so that the model learns the detailed symptoms.\end{tabular}} & \multicolumn{1}{l}{\begin{tabular}[c]{@{}l@{}}-Make the model capable of dialogue in the agriculture field.\\ -Reduce knowledge errors in generated conversations.\end{tabular}} \\ \hline
Format          & Triples (Image, Question, Knowledge)                                                                                                                                                                                     & Image-Knowledge pairs for multi-turn dialogues                                                                                                                                        \\ \hline

\revise{
Geographic Distribution}	 & \revise{Global datasets (e.g., PlantVillage)} &\revise{Focuses on Chinese agricultural scenarios}\\ \hline
\end{tabular}

}
\end{threeparttable}
\scriptsize
\begin{tablenotes}
\item[i] [i] CAAS: Chinese Academy of Agricultural Sciences
\end{tablenotes}
\end{table*}

We evaluate the performance of \sysname based on the following two agricultural benchmarks\cite{Agri_LLaVA}:
\revise{\subsubsection{Chatbot-Bench} A benchmark consisting of 151 dialogues containing 30 pest/disease categories (6 pests, 24 diseases) is created, covering common crops such as strawberries, apples, grapes, potatoes, etc. Among these, 25 categories (4 pests, 21 diseases) are excluded from the training set to evaluate the model's generalization capability on unseen categories. Responses are scored by GPT-4 using a 1–10 scale based on expert annotations assessing relevance, accuracy, and granularity of details.}

\revise{\subsubsection{VQA-Bench} A benchmark containing 2,268 Q\&A pairs is constructed, covering 49 diseases, 50 pests and 6 healthy samples, covering more than 10 crops (e.g., wheat, rice, tomato). Among these, 24 categories (3 pests, 21 diseases) are excluded from the training set. Visual reasoning is evaluated through closed-set accuracy (disease/presence classification) and open-set F1 score (free-form symptom explanation).}

\subsection{Results}

As shown in Table~\ref{tab:result1}, in Agri-Chatbot-Bench, \sysname model goes beyond the TinyLLaVA mini-general model in terms of instruction-following and multi-round conversations. Compared to the existing agricultural proprietary MLLM Agri-LLaVA, through three stages of knowledge distillation, the scale of parameters decreased to approximately 1B, and the GPT-4 score only decreased by 2.9. It performs well in instruction compliance, reaching 51.5\% of the GPT-4 algorithm. 

\begin{table}[]
\caption{The results on Chatbot-Bench}\label{tab:result1}
\centering
\scalebox{1.0}{
\begin{tabular}{ccc}
\hline
Method   & \#Params & Agri-Chatbot Score      \\ \hline
LLaVA                            &$\sim$7B &50.6      \\
TinyLLaVA                        &$\sim$3B &46.8             \\

Agri-LLaVA                       &$\sim$7B &54.4                             \\
\sysname                         &\textbf{$\sim$1B} &51.5  
        
\\ \hline
\end{tabular}
}
\end{table}

\begin{table}[]
\caption{\revise{The results on VQA-Bench}}\label{tab:result2}
\centering
\scalebox{0.88}{
\begin{tabular}{ccc|cc}
\hline
\multirow{2}{*}{Method}   & \multicolumn{2}{c|}{LLM} & \multicolumn{2}{c}{Agri-VQA} \\ \cline{2-5} 
                                &     Name& \#Params          & F1-score (Open)       & Accuracy (Closed)         \\ \hline

Qwen-VL-Chat                     &Qwen & 7B  &30.2 &84.5             \\
Mini-Gemini-2B                  &Gemma &$\sim$2B  &27.3 &81.1             \\
TinyLLaVA                       &Phi2 &$\sim$2.7B  &22.1 &77.5             \\

Agri-LLaVA                      &Vicuna & $\sim$7B  &30.8 &89.3                             \\
\sysname                        &Qwen2.5 &\textbf{$\sim$1B}  &28.7 &85.9   
        
\\ \hline
\end{tabular}
}
\end{table}

\revise{Table~\ref{tab:result2} summarizes the results on VQA-Bench. We use F1-score for open-set questions (e.g., symptom descriptions) with unrestricted answer spaces, and accuracy for closed-set tasks (e.g., category identification) with predefined answer spaces.} As shown in Table~\ref{tab:result2}, in VQA-Bench, compared to existing general models, the \sysname model has a slight advantage in accuracy, but the parameter scale has been significantly reduced. Compared with the existing small-scale general models, it has an obvious advantage in accuracy. Compared with the proprietary model Agri-LLaVA, the accuracy of the closed set task is 85.9\%, which is close to that of Agri-LLaVA, but there is still a performance gap of about 7\% in the open set task.

\begin{comment}
{\color{blue}
The suboptimal F1-score performance stems from two fundamental limitations in our model: (1) class imbalance in the training data, where insufficient samples of rare pests/diseases lead to inadequate feature learning and consequent false negatives during detection, resulting in missed identification of genuinely infected specimens; (2) the inherent challenge of fine-grained discrimination among numerous pest/disease categories with visually similar symptoms, causing misclassification (e.g., disease A being predicted as disease B) in complex scenarios. These issues respectively compromise recall and precision metrics, collectively degrading the F1-score, necessitating targeted improvements in data balancing strategies and feature learning mechanisms.
}
\end{comment}

\begin{figure}
    \centering
    \includegraphics[width=1\linewidth]{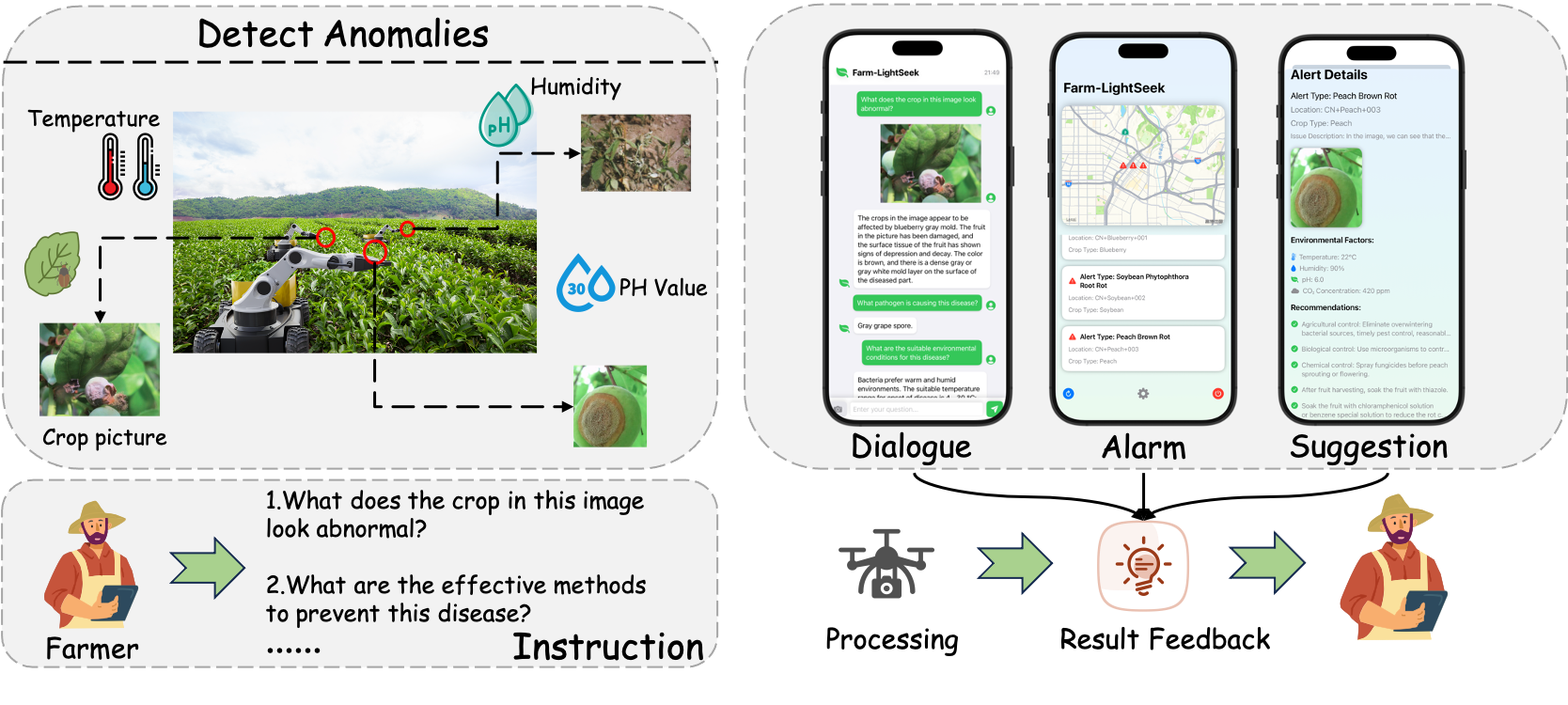}
    \caption{A test sample of \sysname.}
    \label{fig:result picture}
\end{figure}

As shown in \figurename~~\ref{fig:result picture}, our mobile application implementation of \sysname delivers superior anomaly detection capabilities empowered by edge intelligence. By leveraging the direct connection between the mobile device and the edge \revise{nodes}, \sysname bypasses the need for cloud computing servers, significantly reducing service latency. After processing the data, \sysname promptly identifies anomalies and provides comprehensive feedback to the user's device. This feedback includes precise sensor anomaly localization, high-resolution anomaly images, detailed anomaly information, and actionable recommended solutions. The ability to provide such a holistic response not only enhances the system's usability, but also significantly improves the efficiency of anomaly resolution. This result underscores the robustness and practicality of \sysname in real-world edge computing scenarios, where timely and accurate anomaly handling is critical.  

In conclusion, our framework is expected to achieve real-time detection in practical agricultural scenarios, further enhancing its applicability and effectiveness.

\section{Future Directions}
\subsection{Enhancement of Open Set Learning}
While \sysname demonstrates a significant reduction in parameter size, its performance in open set scenarios highlights the inherent limitations of compact models in multi-hop reasoning and the generalization of rare classes. To address these challenges, future research will focus on developing more efficient data augmentation and sampling strategies to ensure that the model comprehensively learns the distinctive features of rare categories during training. Moreover, it is crucial to explore advanced open set learning approaches to improve the model’s ability to generalize when encountering unknown categories. This involves incorporating diffusion models to adaptively learn from limited data and utilizing self-supervised learning methods to enhance feature representation by leveraging unlabeled data. These innovations are expected to significantly improve the robustness and adaptability of \sysname in diverse and dynamic agricultural environments.

\subsection{Security and Robustness}
\revise{The lightweight MLLM deployed on edge nodes in open-field farming environments is vulnerable to adversarial inputs, such as perturbed images or sensor data designed to mislead pest detection or environmental anomaly classification. For instance, subtle noise injected into crop imagery could cause the model to misdiagnose healthy plants as diseased, leading to erroneous irrigation or pesticide recommendations.} To address the dual challenges of privacy and resource constraints in agricultural IoT systems, it is significant to develop edge-optimized homomorphic encryption schemes using lattice-based cryptography, enabling secure fusion of multimodal data without decryption, while minimizing computational overhead. For on-device MLLMs, adaptive differential privacy (DP) mechanisms could dynamically calibrate noise injection during fine-tuning that balances model accuracy against inference risks. Furthermore, the designed verification pipelines should validate the integrity of end-to-end data, from sensor ingestion to actuator output, to prevent cascading failures in agricultural IoT systems. 

\subsection{Edge-centric Autonomous Farming Robotics}
To enable context-aware robotic decision-making in smart agriculture, such as adaptive crop-disease targeting or variable-depth planting, it is crucial to design the swarm robotics coordination protocols for scalable field operations, such as crop density variability. Energy-efficient swarm intelligence solutions are expected to coordinate fleets of edge-powered robots in GPS-denied environments. In addition, embedded causality models could enhance robotic understanding of soil-plant-microbiome interactions, optimizing interventions while minimizing ecological disruption.

\revise{
\subsection{Data-centric Enhancement Strategies for Agricultural Pest Diagnosis Generalization}
The dataset in this study exhibits several limitations. First, it shows imbalanced class distribution, with common pests and diseases overrepresented and rare types underrepresented. Second, the dataset demonstrates geographical bias toward specific regions, lacking global diversity in pest/disease coverage. Third, dependence on GPT-4 for data generation may introduce synthetic biases or descriptive preferences. Additionally, most data originate from public platforms and laboratory environments, resulting in inadequate representation of real-world farmland conditions. Future efforts should focus on: (1) collaborating with multiple agricultural institutions to build geographically balanced datasets; (2) designing a contrastive learning strategy to align features and mitigate potential biases in LLaVA’s agricultural applications and (3) enhancing model generalization in complex field environments through privacy-focused edge-computing solutions (e.g., federated learning with DP) for rare disease diagnostics.}

\subsection{Intelligent Multimodal Energy Management Systems}
An intelligent energy harvesting and management system is crucial to tackle energy sustainability challenges for agricultural devices and edge nodes in long-term field operations. This entails integrating multi-source energy technologies (e.g., solar and vibration harvesting) and refining deep learning models for accurate energy consumption forecasting. By deploying real-time energy tracking and adaptive control mechanisms that dynamically adjust computational precision and communication frequency, the system can maintain a self-balancing equilibrium between energy supply and operational demands. Such a strategy strengthens the resilience of off-grid agricultural tools, enabling energy-efficient decision-making and supporting sustainable performance in remote rural environments.

\section{Conclusion}
In this article, we introduce \sysname, a lightweight multimodal intelligent data analytics framework for agricultural IoT. This framework addresses the key challenges of high data heterogeneity, high-latency data processing, and costly deployment in traditional agricultural monitoring and decision-making systems by leveraging knowledge distillation and multimodal data fusion techniques.

\sysname integrates multi-source information such as agricultural text data and high-resolution images, and realizes end-to-end pest classification, environmental anomaly detection and dynamic decision generation based on lightweight large model architecture. The experimental results show that \sysname has a GPT-4 score of 51.5 in the agricultural dialogue task and 85.9\% accuracy in the closed set visual Q\&A, which verifies its reliability in the pest classification task. Although there is still room for optimization in the reasoning ability of open set, the acceleration of reasoning and the reduction of memory footprint make it possible to achieve fast real-time response on edge nodes and meet the needs of real-time decision-making in the field.

\bibliographystyle{ieeetr}
\bibliography{ref.bib}

@inproceedings{LLaVa,
 author = {Liu, Haotian and Li, Chunyuan and Wu, Qingyang and Lee, Yong Jae},
 booktitle = {Proceedings of the Advances in Neural Information Processing Systems (NeurIPS)},
 pages = {34892--34916},
 title = {Visual Instruction Tuning},
 url = {https://proceedings.neurips.cc/paper_files/paper/2023/file/6dcf277ea32ce3288914faf369fe6de0-Paper-Conference.pdf},
 volume = {36},
 year = {2023}
}

@article{LLaVA-KD,
      title={{LLaVA-KD}: A Framework of Distilling Multimodal Large Language Models}, 
      author={Yuxuan Cai and Jiangning Zhang and Haoyang He and Xinwei He and Ao Tong and Zhenye Gan and Chengjie Wang and Xiang Bai},
      year={2024},
      journal={arXiv preprint arXiv:2410.16236},
      eprint={2410.16236},
      archivePrefix={arXiv},
      primaryClass={cs.CV},
      url={https://arxiv.org/abs/2410.16236}, 
}

@article{TinyLLaVA,
      title={{TinyLLaVA}: A Framework of Small-scale Large Multimodal Models}, 
      author={Baichuan Zhou and Ying Hu and Xi Weng and Junlong Jia and Jie Luo and Xien Liu and Ji Wu and Lei Huang},
      year={2024},
      journal={arXiv preprint arXiv:2402.14289},
      eprint={2402.14289},
      archivePrefix={arXiv},
      primaryClass={cs.LG},
      url={https://arxiv.org/abs/2402.14289}, 
}

@ARTICLE{PaganoIoT23,
  author={Pagano, Antonino and Croce, Daniele and Tinnirello, Ilenia and Vitale, Gianpaolo},
  journal={IEEE Internet of Things Journal}, 
  title={A Survey on {LoRa} for Smart Agriculture: Current Trends and Future Perspectives}, 
  year={2023},
  volume={10},
  number={4},
  pages={3664-3679},
  doi={10.1109/JIOT.2022.3230505}}

@article{Agri_LLaVA,
      title={{Agri-LLaVA}: Knowledge-Infused Large Multimodal Assistant on Agricultural Pests and Diseases}, 
      author={Liqiong Wang and Teng Jin and Jinyu Yang and Ales Leonardis and Fangyi Wang and Feng Zheng},
      year={2024},
      journal={arXiv preprint arXiv:2412.02158},
      eprint={2412.02158},
      archivePrefix={arXiv},
      primaryClass={cs.CV},
      url={https://arxiv.org/abs/2412.02158}, 
}

@inproceedings{GEOCLIP,
  title={Geoclip: Clip-inspired alignment between locations and images for effective worldwide geo-localization},
  author={Vivanco Cepeda, Vicente and Nayak, Gaurav Kumar and Shah, Mubarak},
  booktitle={Proceedings of the Advances in Neural Information Processing Systems (NeurIPS)},
  volume={36},
  pages={8690--8701},
  year={2023}
}

@InProceedings{FNKD,
author="Xu, Kunran
and Rui, Lai
and Li, Yishi
and Gu, Lin",
title="Feature Normalized Knowledge Distillation for Image Classification",
booktitle="Proceedings of the European Conference on Computer Vision (ECCV)",
year="2020",
pages="664--680",

}

@InProceedings{DGKD,
    author    = {Son, Wonchul and Na, Jaemin and Choi, Junyong and Hwang, Wonjun},
    title     = {Densely Guided Knowledge Distillation Using Multiple Teacher Assistants},
    booktitle = {Proceedings of the IEEE/CVF International Conference on Computer Vision (ICCV)},

    year      = {2021},
    pages     = {9395-9404}
}

@inproceedings{minillm,
      title={{MiniLLM}: Knowledge Distillation of Large Language Models}, 
      author={Yuxian Gu and Li Dong and Furu Wei and Minlie Huang},
      year={2024},
      eprint={2306.08543},
      booktitle={Proceedings of the International Conference on Learning Representations (ICLR)} 
}

@inproceedings{distillm,
      title={{DistiLLM}: Towards Streamlined Distillation for Large Language Models}, 
      author={Jongwoo Ko and Sungnyun Kim and Tianyi Chen and Se-Young Yun},
      year={2024},
      eprint={2402.03898},
booktitle={Proceedings of the International Conference on Machine Learning (ICLR)},
 no = {997}, 
pages = {24872 - 24895}      
}

@inproceedings{tinyllm,
       author = {{Tian}, Yijun and {Han}, Yikun and {Chen}, Xiusi and {Wang}, Wei and {Chawla}, Nitesh V.},
        title = "{Beyond Answers: Transferring Reasoning Capabilities to Smaller LLMs Using Multi-Teacher Knowledge Distillation}",
booktitle = {International Conference on Web Search and Data Mining (WSDM)},
year = 2024,

}

@inproceedings{llavamod,
      title={{LLaVA-MoD}: Making {LLaVA} Tiny via {MoE} Knowledge Distillation}, 
      author={Fangxun Shu and Yue Liao and Le Zhuo and Chenning Xu and Lei Zhang and Guanghao Zhang and Haonan Shi and Long Chen and Tao Zhong and Wanggui He and Siming Fu and Haoyuan Li and Bolin Li and Zhelun Yu and Si Liu and Hongsheng Li and Hao Jiang},
      year={2025},
      eprint={2408.15881},
      booktitle = {Proceedings of the International Conference on Learning Representations (ICLR) }
      
}

@article{MUHAMMED2024103905,
title = {Artificial Intelligence of Things {(AIoT)} for smart agriculture: A review of architectures, technologies and solutions},
journal = {Journal of Network and Computer Applications},
volume = {228},
pages = {103905},
year = {2024},
issn = {1084-8045},
author = {Dalhatu Muhammed and Ehsan Ahvar and Shohreh Ahvar and Maria Trocan and Marie-José Montpetit and Reza Ehsani}
}

@ARTICLE{LiTPAMI25,
  author={Li, Yunxin and Jiang, Shenyuan and Hu, Baotian and Wang, Longyue and Zhong, Wanqi and Luo, Wenhan and Ma, Lin and Zhang, Min},
  journal={IEEE Transactions on Pattern Analysis and Machine Intelligence}, 
  title={{Uni-MoE}: Scaling Unified Multimodal {LLMs} with {Mixture of Experts}}, 
  year={2025},
  volume={},
  number={},
  pages={1-15},
  doi={10.1109/TPAMI.2025.3532688}}

@ARTICLE{XuIEEE21,
  author={Xu, Dianlei and Li, Tong and Li, Yong and Su, Xiang and Tarkoma, Sasu and Jiang, Tao and Crowcroft, Jon and Hui, Pan},
  journal={Proceedings of the IEEE}, 
  title={Edge Intelligence: Empowering Intelligence to the Edge of Network}, 
  year={2021},
  volume={109},
  number={11},
  pages={1778-1837},
  doi={10.1109/JPROC.2021.3119950}}

\end{document}